\pdfoutput=1

\documentclass[11pt]{article}

\usepackage[dvipsnames]{xcolor}
\usepackage{tcolorbox}
\usepackage{authblk}
\usepackage{NACL2025}
\usepackage{hyperref}
\usepackage{times}
\usepackage{latexsym}
\usepackage{booktabs}
\usepackage{placeins}
\usepackage[T1]{fontenc}

\usepackage[utf8]{inputenc}

\usepackage{microtype}

\usepackage{inconsolata}
\usepackage{array} 

\usepackage{graphicx}
\usepackage{todonotes}
\usepackage{amsmath}
\usepackage{multirow}
\usepackage{multicol}
\usepackage{caption}
\usepackage{subcaption}
\usepackage{float}

\usepackage[labelfont=bf]{caption}
\usepackage{adjustbox}

\usepackage{natbib}

\title{Thought2Text: Text Generation from EEG Signal using Large Language Models (LLMs)}  

\author{ \textbf{\ Abhijit Mishra$^*$,\ Shreya Shukla$^*$, \ Jose Torres, \ Jacek Gwizdka,\ Shounak Roychowdhury} \\
        School of Information, University of Texas at Austin\\
        \{abhijitmishra, shreya.shukla, jtorres1221, jacekg, shounak.roychowdhury\}@utexas.edu}

\begin{document}
\maketitle

\def\thefootnote{*}\footnotetext{Equal Contribution}\def\thefootnote{\arabic{footnote}}

\begin{abstract}
Decoding and expressing brain activity  in a comprehensible form is a challenging frontier in AI. This paper presents \textit{Thought2Text}, which uses instruction-tuned Large Language Models (LLMs) fine-tuned with EEG data to achieve this goal. The approach involves three stages: (1) training an EEG encoder for visual feature extraction, (2) fine-tuning LLMs on image and text data, enabling multimodal description generation, and (3) further fine-tuning on EEG embeddings to generate text directly from EEG during inference. Experiments on a public EEG dataset collected for six subjects with image stimuli and text captions demonstrate the efficacy of multimodal LLMs (\textsc{LLaMA-v3}, \textsc{Mistral-v0.3}, \textsc{Qwen2.5}), validated using traditional language generation evaluation metrics, as well as \textit{fluency} and \textit{adequacy} measures. This approach marks a significant advancement towards portable, low-cost "thoughts-to-text" technology with potential applications in both neuroscience and natural language processing.
\end{abstract}

\section{Introduction}
\label{sec:intro}
Brain-Computer Interface (BCI) systems, combined with portable and wearable noninvasive \textit{Electroencephalographic} (EEG) devices, enable direct interfacing between the brain and external devices \cite{he2015noninvasive}. Advances in generating images and natural language using EEG \cite{Speier_2016, benchetrit2023brain, defossez2023decoding} hold promise for developing BCIs in various domains, including assistive communication (\textit{e.g.,} for ALS and stroke patients), mixed reality (AR/VR) experience enhancement, mental health diagnosis, and gaming. Recent strides in NLP driven by powerful Large Language Models (LLMs) such as OpenAI GPT-4. \cite{achiam2023gpt}, Google Gemini \cite{team2023gemini}, Meta-LLaMA \cite{touvron2023llama}, Mistral \cite{jiang2023mistral}, and Microsoft Phi \cite{gunasekar2023textbooks} have enabled multi-modal integration, facilitating language generation from images \cite{liu2024visual} and speech \cite{fathullah2024prompting}. Our research focuses on a multi-modal solution to decode brain signals directly into text, using EEG signals. We choose EEG because of its affordability compared to mainstream alternatives such as Functional Magnetic Resonance Imaging (fMRI) \cite{tang2023semantic}, which are costlier and require complex setup. For generating text, we leverage LLMs, which enable flexible, high-quality text generation across various modalities, such as images \cite{liu2024visual}, audio \cite{rubenstein2023audiopalm, fathullah2024prompting}, and more and often outperform current open vocabulary task-specific methods \cite{shi2024llmformer}.

Our approach for generating textual descriptions from EEG signals involves three key steps: (a) capturing language-agnostic EEG signals via visual stimuli, (b) encoding these signals into embeddings using a deep multichannel neural encoder, and (c) fine-tuning language models by projecting image and EEG embeddings into a token embedding space to generate responses. These responses are compared with gold standard image descriptions to compute the training loss. During inference, only EEG signals and a generic textual prompt are used as inputs to the LLMs to generate responses. Our method requires images, EEG data and descriptions for training, while inference is bimodal, using only EEG to generate text.

For experiments, we use a public 128-channel EEG dataset from six participants viewing visual stimuli. The image descriptions are generated by GPT-4-Omni \cite{achiam2023gpt} and quality-checked by human annotators, providing the text modality necessary to build EEG-to-text generation systems. Although the goal is the generation of text from EEG, we use a dataset with visual stimuli for their language-agnostic nature, avoiding the potential complexities associated with reading and language processing (see Section \ref{sec:dataset} for details). We fine-tune large language models (LLMs) using these descriptions, leveraging pre-trained instruction-based language models such as \textsc{Mistral-v3} \cite{jiang2023mistral}, \textsc{LLaMA-v3} \cite{touvron2023llama}, and \textsc{Qwen2.5} \cite{bai2023qwen}. Evaluation with standard generation metrics \cite{sharma2017nlgeval} and GPT-4-based assessments confirmed the effectiveness of our approach.

Our paper's key contributions include:
\begin{itemize}
    \item Integration of brain signals with instruction-tuned LLMs.
    \item Fine-tuning models on EEG signals captured for visual stimuli, leveraging its language-agnostic nature to enhance LLM interaction. 
    \item Validation of the efficacy of the model in a popular public dataset that contains EEG signals captured using affordable devices. 
\end{itemize}

The code and a link to the processed dataset can be found at \url{https://github.com/abhijitmishra/Thought2Text}. 

\section{Related Work}
Integrating behavioral signals such as eye movement and brain signals into NLP and computer vision tasks \cite{mishra2018cognitively,sharma2024emerging} has seen significant progress. Key datasets include \textit{ZuCo 2.0} \cite{hollenstein2019zuco}, which captures EEG and eye-gaze during natural language reading, and \textit{MOABB} \cite{Jayaram_2018}, offering over 120,000 EEG samples from 400+ subjects from various BCI tasks such as motor imagery, visual evoked potentials, and cognitive load. Datasets such as \textit{MindBigData} \cite{vivancos2022mindbigdata} and \textit{CVPR2017} \cite{spampinato2017deep} provide substantial EEG data collected from participants' responses to handwritten and open vocabulary object-based image stimuli respectively. These datasets have facilitated research on the classification of EEG data \cite{spampinato2017deep,palazzo2020decoding,khaleghi2023salient} and the generation of images from EEG signals using GANs \cite{goodfellow2020generative} and latent diffusion models \cite{rombach2022high,bai2023dreamdiffusion,lan2023seeing,tirupattur2018thoughtviz}. 

Additionally, multimodal datasets like \textit{The Alice Dataset} \cite{bhattasali2020alice}, which includes EEG and fMRI recordings from participants listening to a story, provide two measurable modalities: audio stimulus and the corresponding text. Another recent multimodal dataset, \textit{EIT-1M} \cite{zheng2024eit}, contains one million EEG-Image-Text pairs, collected as participants viewed visual-textual stimuli. At the time of writing, a partial version of the \textit{EIT-1M} dataset was released, containing data for only one subject.

Generating language from EEG signals remains an elusive challenge. The most closely related work is \textit{EEG2TEXT} \cite{liu2024eeg2text}, which utilizes EEG pretraining and a multi-view transformer to decode EEG signals into text. Another approach similar to ours, using multiple modalities, is presented in \cite{ikegawa2024text}, where intracranial EEG (iEEG) signals were recorded from patients watching videos, and each video frame was used to generate images and text using CLIP vision-language model \cite{radford2021learning}. Unlike our approach, this study involved implanting electrodes in patients. Furthermore, these works do not leverage large language models (LLMs) with prompt engineering for generating prompt-specific responses. We believe our method of fine-tuning LLMs using non-invasive EEG input is the first of its kind.
\section{Dataset and the need for Visual Stimuli}
\label{sec:dataset}
Building a system that generates text from neural activity naturally requires a dataset of paired $<eeg,~text>$ examples. However, using textual stimuli presents inherent challenges. Reading is a learned skill that requires the decoding of symbols into sounds and meanings, syntactic parsing, and sequential integration in time. In contrast, visual perception is more innate, natural, and image processing is more parallel \cite{dehaene_reading_2009, townsend_serial_1990}.  Furthermore, using EEG data collected on textual stimuli introduces additional complexities of language processing, such as determining brain activity windows for specific words, managing retention of word context post-onset, and managing the overlap of contexts when words are shown in different time frames \cite{wehbe2014aligning, murphy2022decoding}. Additionally, vocabulary size presents a challenge: while EEG-to-text systems perform well in closed-vocabulary settings, open-vocabulary decoding becomes inefficient as vocabulary size increases \cite{martin2018decoding, wang2022open, liu2024eeg2text}. The core challenge in thought-to-text systems thus lies in two key aspects: \textit{(a) harnessing language-agnostic neural signals with minimal interference from linguistic processing} and \textit{(b) using these signals to generate text in a target language}, potentially by specifying an instruction or prompt.

This leads us to experiment with datasets where EEG signals are recorded in response to visual stimuli rather than textual input. For \textit{(a)}, visual stimuli provide several advantages: By relying on images, it circumvents the complexities of language processing and also elicits brain responses to salient image features, making them more suited to capture neural activity in a language-agnostic manner. For \textit{(b)}, generating text from visual stimuli-evoked EEG remains challenging, as most EEG datasets collected with visual stimuli (e.g., \textit{CVPR2017} dataset \cite{spampinato2017deep}, \textit{MOABB} \cite{Jayaram_2018}, \textit{MindBigData} \cite{vivancos2022mindbigdata}) lack an associated text component, which is essential for evaluating whether the generated text from EEG signals accurately captures the perceived salient features of the images. To address this, robust image captioning tools such as GPT-4 Omni \cite{achiam2023gpt} can be employed to generate captions that effectively capture the salient imformation of visual stimuli, thus creating a tri-modal dataset with $<eeg,~text,~image>$ tuples. Text generation can further be guided by specifying a prompt or instruction to tailor the output to a particular language or context. This approach enriches the dataset and provides a more comprehensive foundation for thought-to-text research.

\begin{table}[t]
    \centering
    \footnotesize
    \begin{tabular}{l  c}
        \toprule
         \textbf{Annotator}& \textbf{Percentage of Correct Captions}\\ 
         \midrule
         Annotator 1	& 98\%\\ 
         \midrule
         Annotator 2	& 96\%\\
         \midrule
         Agreement Score &93\%\\
         \bottomrule
    \end{tabular}
    \caption{GPT-4 Captions Validation Results from Amazon Mechanical Turk }
    \label{tab:annotator_data}
\end{table}

For our experiments, we utilize the \textit{CVPR2017} dataset \cite{spampinato2017deep}, which contains preprocessed EEG data from six participants, each viewing 50 images across 40 diverse object categories\footnote{we use the terms \textit{object} and \textit{class} interchangeably to refer to an object category.} (such as vehicles, musical instruments, \textit{etc}). Each EEG recording, corresponding to one participant and one visual stimulus, consists of 128 channels recorded for 0.5 seconds at a sampling rate of 1 kHz. Data are represented as a 128×N matrix, where N, approximately 500, represents the number of samples per channel in each segment. According to \newcite{spampinato2017deep}, the EEG signals were pre-processed by first applying a second-order Butterworth bandpass filter between 5 Hz and 95 Hz and a notch filter at 50 Hz to remove power line noise. In addition, since the exact duration of the EEG signals can vary, the first 20 samples (20 ms) were discarded to reduce interference from the previous image, and then standardizing the signal to a common length of 440 samples, accounting for segments with N < 500. The dataset provides pre-filtered signals across three frequency ranges: 14-70Hz, 5-95Hz, and 55-95Hz. In line with previous research \cite{palazzo2020decoding}, we selected the 55-95 Hz range, as it has shown the most reliable results. We used the training, validation, and test split from the original paper for overall and subject-wise analysis. 

Since the \textit{CVPR2017} dataset lacks textual descriptions, we used GPT-4 to generate brief one-line captions for each image as they are efficient in capturing salient information without introducing extraneous details. Here, the aim is to map EEG data to concise captions to simplify the alignment task, making it feasible despite the noise and variability inherent in EEG signals. To ensure quality, we validated these auto-generated captions using human annotators from \textit{Amazon Mechanical Turk}, who rated their fluency and adequacy on a binary scale. Table \ref{tab:annotator_data} shows the validation results, including annotator agreement on acceptable captions, highlighting the reliability of the generated captions. We release the descriptions under the same license as the original data. 
\begin{figure*}[t]
     \centering
    {\includegraphics[width=\textwidth]{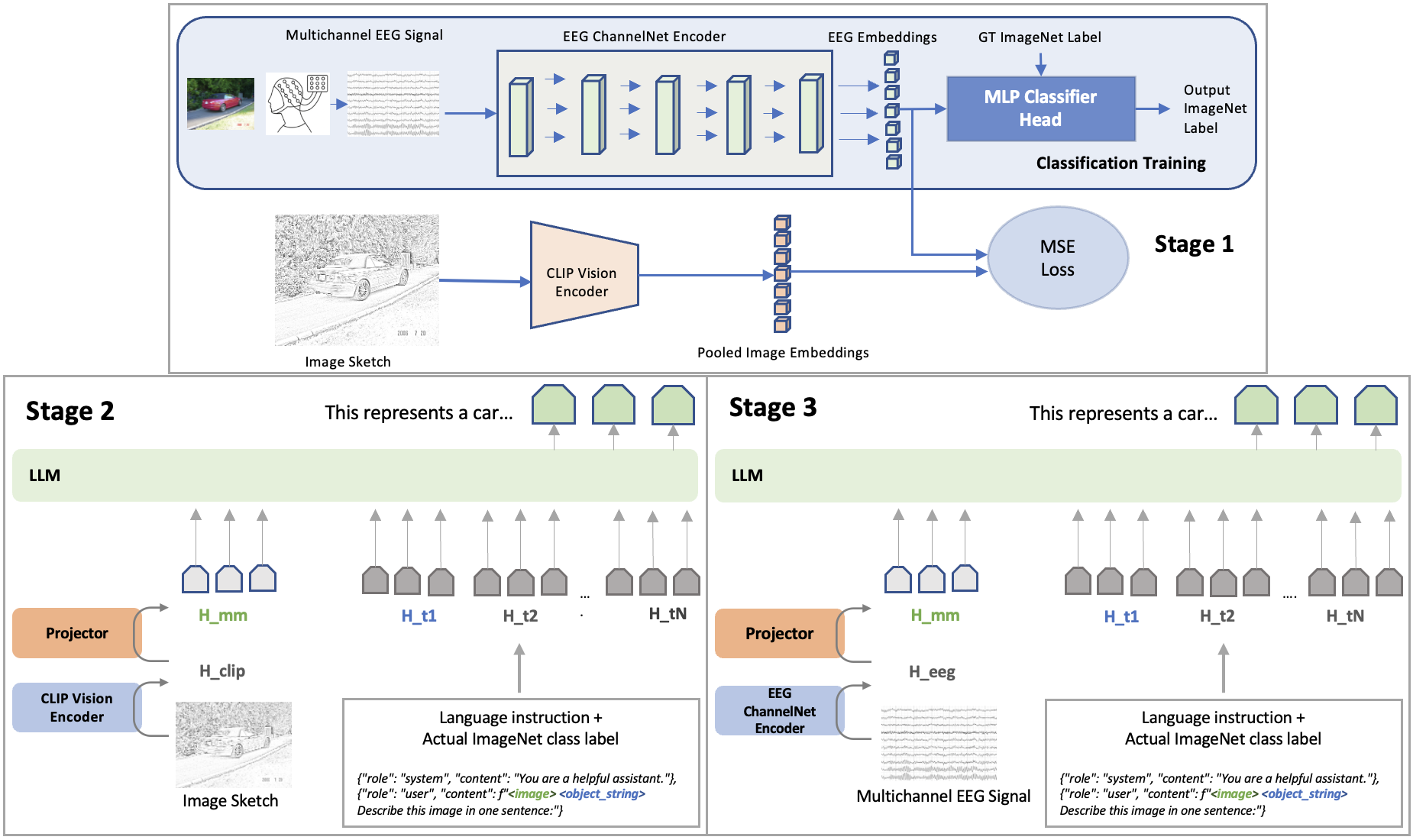}}
   \caption{Multi-stage training process for \textit{Thought2Text}. \textit{\textbf{Stage 1:}} EEG ChannelNet encoder is trained using MSE loss (aligning EEG embeddings with CLIP model image embeddings) and CE loss (for EEG classification). \textit{\textbf{Stage 2:}} LLMs are fine-tuned using image descriptions and object labels, with only the projector trained while the LLM and CLIP encoder remain frozen. \textit{\textbf{Stage 3:}} Similar to stage 2, but the projector is trained with EEG embeddings. LLM and EEG encoder remain frozen.}
	\label{fig:overall}
\end{figure*}
\section{Method}
\label{sec:method}
Our approach uses a three-stage process to generate coherent text from EEG signals. The overall workflow is illustrated in Figure \ref{fig:overall}.

\subsection{Stage1: Training EEG Encoder for Embedding Extraction}
The first stage of our approach focuses on developing an encoder that extracts meaningful embeddings ($H_{eeg}$) from multi-channel EEG signals. Since the target thoughts are short and pertain to the most salient features of an image, we design the encoder with two objectives: (a) aligning the EEG embeddings with those derived from image stimuli using a pretrained visual encoder, and (b) predicting the most salient object (\textit{e.g.,} piano) from the EEG embeddings. 

As shown in Figure \ref{fig:overall}, we use a multichannel EEG encoder inspired by \textit{ChannelNet} \cite{palazzo2020decoding}, a deep convolutional neural network model, to convert EEG signals into multidimensional embeddings ($H_{eeg}$). These embeddings are further processed by an MLP classifier to predict object labels ($y_{obj}$), which correspond to objects present in the image stimuli. The label-set matches ImageNet’s labels, available as part of the dataset, representing the most salient object in each image.

The model is trained by minimizing two losses: (A) a categorical cross-entropy loss ($CE$) between the predicted and ground-truth object labels, and (B) a mean squared error ($MSE$) between the EEG embeddings ($H_{eeg}$) and pooled image embeddings ($H_{clip}$) from a pretrained CLIP model \cite{radford2021learning}, which captures semantically rich image representations. CLIP’s embeddings offer robust transfer learning capabilities, making them ideal for aligning with EEG data. Performance metrics for EEG-to-image classification using trained MLP Classifier 

To better align EEG embeddings with visual stimuli, we simplify images by removing non-central details like color, converting them into sketches using techniques such as \textit{Gaussian Blur} and \textit{Canny} filters. Although this step is empirical and optional, using sketches help focus the alignment on the core features of the object as discussed in Fine-Grained Sketch-Based Image Retrieval \cite{luo2023review} and Interactive Sketch Question Answering \cite{lei2024emergent}.

The general loss function ($\mathcal{L}$) balances $MSE$ and $CE$, weighted by a hyper-parameter $\alpha$ (set to 0.5):

\begin{align}
\mathcal{L} &= (1 - \alpha) \cdot \text{MSE}(\mathbf{H}_{eeg}, \mathbf{H}_{clip}) \nonumber \\ 
&\quad + \alpha \cdot \text{CE}(\mathbf{y}_{obj}, \mathbf{\hat{y}}_{obj})
\end{align}

We predict both $H_{eeg}$ and $y_{obj}$ for three reasons: (a) aligning EEG embeddings with image embeddings allows us to leverage multimodal vision-language models later, adapting pretrained models for EEG-based text generation, (b) joint optimization ensures the embeddings emphasize salient objects in the images, and (c) object labels, combined with EEG embeddings, can later be fed into multimodal language models to guide more accurate generation. Given the noisy nature of EEG signals, including object labels in the prompts helps keep the model grounded in the salient object and reduces the likelihood of hallucinations.

\subsection{Stage2: Priming LLMs with Image Embeddings}
\label{subsec:priming}
To enable LLMs to process multimodal inputs, such as EEG and visual embeddings, we designed a projector inspired by recent advancements in vision-language models \cite{zhang2024vision} and is a fully connected feed-forward layer. Since LLMs are inherently text-based and cannot natively accept non-text embeddings, the projector transforms embeddings from the vision and EEG models into the token embedding space of the LLM. This ensures that the projected embeddings have the same dimensionality as the LLM’s token embeddings.

The projector is a simple feed-forward layer that maps the embeddings to the LLM token space. These transformed embeddings are then concatenated with token embeddings extracted from the input prompt, allowing the LLM to process both text and external modality embeddings seamlessly. In our main setting, the input prompt is structured as follows:

\begin{figure}[H]
\centering
\begin{tcolorbox}[colback=gray!10, colframe=gray!30, coltitle=black,
    fonttitle=\bfseries, left=0.1cm, right=0.1cm, bottom=-1mm-0.5mm]
\{"role": "system", "content": "You are a helpful assistant."\},\\
\{"role": "user", "content": "<image> <object\_string> Describe this in one sentence:"\},
\tcbsubtitle[halign=center]{Input prompt}
\end{tcolorbox}
\end{figure}

In stage 2, we first embed the tokens from the extended prompt, including the object labels, using the LLM’s token embedding layer. For this, text tokens are input as token IDs that were converted to dense vectors via a lookup in the LLM's\textit{ embed\_tokens} layer. The token embeddings, $H_{t_1}, H_{t_2}, ..., H_{t_N}$, are then augmented with multimodal embeddings. Specifically, the embedding for the $<image>$ token is replaced by the projected multimodal embedding $H_{mm}$, and the token-embeddings for the $<object\_string>$ is replaced by the embeddings of the ground-truth object label. The multimodal embedding $H_{mm}$ is computed by projecting pooled embeddings from CLIP, $H_{clip}$, into the LLM’s token embedding space using the following transformation:

\begin{equation}
\mathbf{H}_{mm} = \mathbf{W}_{mm} \cdot \mathbf{H}_{clip} + \mathbf{b}_{mm}
\end{equation}

Here, $\mathbf{W}_{mm}$ and $\mathbf{b}_{mm}$ represent the projector parameters. Once $H_{mm}$ is computed, it is prepended to the token embeddings from the input prompt, enabling the LLM to process multimodal information. 

During training, labels are created by right-shifting the tokens in the prompt, aligning them with the ground-truth image descriptions. Special tokens such as beginning of sentences ($BoS$), end of sentences ($EoS$) padding tokens ($PAD$), system and user and assistant message indicator tokens are used considering LLM specific tokenizers. The input prompts are also converted into LLM specific chat templates. The LLM is then fine-tuned using a standard cross-entropy loss between the predicted tokens and the actual tokens from the ground-truth descriptions.
\subsection{Stage3: Tuning LLMs with EEG Embeddings}
This stage closely resembles stage 2, with the distinction that instead of $H_{clip}$, we utilize $ H_{eeg} $, extracted from the EEG encoder trained in stage 1, to compute multimodal embeddings $ H_{mm} $. During this stage, the projector parameters \( \mathbf{W}_{mm} \) and \( \mathbf{b}_{mm} \) are further tuned. We would like to highlight that throughout this and the previous stage, only the projector is trained, while the LLM and EEG encoders remain frozen to mitigate parameter instability caused by EEG noise. 

\subsection{Inference}
During inference, the EEG ChannelNet encoder processes EEG signals to generate EEG embeddings \( H_{eeg} \). The multimodal projector, trained in stage 3, transforms \( H_{eeg} \) into multimodal embeddings \( H_{mm} \). \( H_{eeg} \) is also fed into the MLP Classifier, trained in stage 1, to predict the object label. This predicted label is appended to a generic input prompt, similar to the one mentioned in Section \ref{subsec:priming}, and the token embeddings are computed for the combined input. Finally, the token embeddings from the projector and the language prompt are concatenated and fed into the LLM to generate descriptions. Notably, no images are used during inference, making the process strictly bimodal.
\section{Experimental Details}
In this section we highlight the dataset, model details, and evaluation procedure. 
\subsection{Dataset} We utilize the open-source \textit{CVPR2017} dataset \cite{spampinato2017deep}, licensed for academic research, featuring EEG signals from six subjects viewing 50 images across 40 ImageNet classes \cite{deng2009imagenet}, totaling 2000 images. The data is split into training ($7959$), evaluation ($1994$), and test ($1987$) examples. Additional details can be found under Section \ref{sec:dataset}. 

\subsection{Model Details} 
We use \textit{ChannelNet} \cite{palazzo2020decoding} for the EEG encoder, modifying the final linear layer to produce 512-dimensional embeddings to match the output of the CLIP vision encoder (\textit{openai/clip-vit-base-patch32}). The EEG encoder is trained with a batch size of 16 for 100 epochs, using the AdamW optimizer and a learning rate of $1e^{-4}$. For fine-tuning the LLM, we use a similar setup: a batch size of 16, training for 5 epochs per stage, and employing gradient accumulation and checkpointing. The learning rate for the LLM fine-tuning is kept at $2e^{-5}$. All implementations are carried out using \textit{PyTorch} and Huggingface's \textit{transformers} library.

We evaluate three LLMs: \textsc{LLaMA-v3} (\textit{meta-llama/Meta-Llama-3-8B-Instruct}), \textsc{Mistral-v0.3} (\textit{mistralai/Mistral-7B-Instruct-v0.3}), and \textsc{Qwen2.5-7B} (\textit{Qwen/Qwen2.5-7B-Instruct}), selected for their efficiency on consumer-grade GPUs such as the NVIDIA RTX 4060Ti. The multimodal embedding $H_{mm}$ is projected to match the token embedding dimensions required by each LLM. All models are permissively licensed for academic research, and training takes approximately 8 GPU hours per LLM training cycle. During inference, we use a batch size of 1, with generation parameters such as \texttt{top\_k}, \texttt{top\_p}, and temperature set to their default values for each LLM.

To demonstrate the effectiveness of our approach, we compare it against the following baselines: (a) \textbf{$\mathrm{ONLY\_OBJ}$}: where LLMs generate a description based solely on the predicted object without any additional input (e.g., if the object is "car," the LLM generates a description of the word "car" following the prompt in Section 4.2); (b) \textbf{$\mathrm{ONLY\_OBJ+RAND\_EMB}$}: where we pass a random embedding alongside the predicted object labels to the LLMs; (c) \textbf{$\mathrm{NO\_STAGE2}$}: where the priming step described in Section \ref{sec:method} is skipped; and (d) \textbf{$\mathrm{ONLY\_EEG}$}: where only the EEG embeddings from Stage 1 are used as input, ignoring the object labels. Our proposed Thought2Text solution, which incorporates all stages and all inputs, is labeled \textbf{ALL} in the experiments.

\subsection{Evaluation} We use standard NLG metrics such as BLEU, METEOR, ROUGE \cite{sharma2017nlgeval}, and BERTScore \cite{zhang2019bertscore}. Furthermore, we assess the quality of the generation using GPT-4, following \cite{liu2024visual}.GPT-4 measures two aspects: \textit{fluency}, for grammar , and \textit{adequacy}, for accuracy in conveying meaning, both rated on a scale of 1-5, with 5 denoting the highest quality.

\begin{table*}[t]
    \centering
    \footnotesize
    \begin{tabular}{lcccccccl|ll} 
         \toprule
         \textbf{LLM} & \multicolumn{3}{c}{\textbf{ROUGE-N}} & \textbf{ROUGE-L}& \multicolumn{2}{c}{\textbf{BLEU-N}} & \textbf{MET-}&    \textbf{BERT}& \textbf{GPT-4}&\textbf{GPT-4}\\ 
          &  N=1&  \multicolumn{2}{c}{N=2}&&  N=1&  N=4& \textbf{EOR} &    \textbf{Score}&\textbf{Flu.}&\textbf{Ade.}\\ 
         \midrule
         $\mathrm{LLaMA3\textrm{-}8B_{ONLY\_OBJ}}$ & 9.8 & \multicolumn{2}{c}{1.5} & 8.5 & 7.3 & 1.3 & 12.6 & 0.84 & 3.44 & 1.30 \\
         $\mathrm{LLaMA3\textrm{-}8B_{OBJ+RAND\_EMB}}$ & 3.8 & \multicolumn{2}{c}{0.4} & 3.3 & 2.8 & 0.4 & 5.9 & 0.84 & 4.72 & 1.08 \\
         $\mathrm{LLaMa3\textrm{-}8B_{ONLY\_EEG}}$ & 28.9 & \multicolumn{2}{c}{7.3} & 26.2 & 24.1 & 5.2 & 23.7 & 0.89 & 4.80 & 1.49 \\
         $\mathrm{LLaMA3\textrm{-}8B_{NO\_STAGE2}}$ & 26.9 & \multicolumn{2}{c}{6.1} & 23.9 & 22.6 & 4.3 & 23.7 & 0.88 & 4.83 & 1.41 \\
         $\mathrm{LLaMA3\textrm{-}8B_{ALL}}$ & \textbf{30.0} & \multicolumn{2}{c}{\textbf{8.1}} & \textbf{26.6} & \textbf{25.5} & \textbf{5.5} & \textbf{26.3} & \textbf{0.89} & \textbf{4.82} & \textbf{1.58} \\
         \midrule
         $\mathrm{Mistral\textrm{-}7B\textrm{-}v0.3_{ONLY\_OBJ}}$ & 17.6 & \multicolumn{2}{c}{3.4} & 14.8 & 14.5 & 2.5 & 23.2 & 0.86 & 4.46 & 1.52 \\
         $\mathrm{Mistral\textrm{-}7B\textrm{-}v0.3_{OBJ+RAND\_EMB}}$ & 17.9 & \multicolumn{2}{c}{3.6} & 15.1 & 15.7 & 2.9 & 22.8 & 0.87 & 4.89 & 1.55 \\
         $\mathrm{Mistral\textrm{-}7B\textrm{-}v0.3_{ONLY\_EEG}}$ & 26.7 & \multicolumn{2}{c}{5.3} & 23.5 & 23.3 & 4.2 & 22.0 & 0.88 & 4.82 & 1.25 \\
         $\mathrm{Mistral\textrm{-}7B\textrm{-}v0.3_{NO\_STAGE2}}$ & 29.2 & \multicolumn{2}{c}{7.3} & 26.5 & 24.1 & 5.0 & 24.0 & 0.89 & 4.77 & 1.60 \\
         $\mathrm{Mistral\textrm{-}7B\textrm{-}v0.3_{ALL}}$ & \textbf{30.6} & \multicolumn{2}{c}{\textbf{8.8}} & \textbf{28.0} & \textbf{26.0} & \textbf{6.1} & \textbf{26.2} & \textbf{0.89} & \textbf{4.79} & \textbf{1.65} \\
         \midrule
         $\mathrm{Qwen2.5\textrm{-}7B_{ONLY\_OBJ}}$ & 17.6 & \multicolumn{2}{c}{2.8} & 14.5 & 14.8 & 2.4 & 21.0 & 0.85 & 3.91 & 1.47 \\
         $\mathrm{Qwen2.5\textrm{-}7B_{OBJ+RAND\_EMB}}$ & 1.7 & \multicolumn{2}{c}{0.1} & 1.6 & 1.3 & 0.3 & 6.4 & 0.84 & 4.73 & 1.01 \\
         $\mathrm{Qwen2.5\textrm{-}7B_{ONLY\_EEG}}$ & 25.2 & \multicolumn{2}{c}{3.6} & 21.5 & 21.9 & 3.2 & 20.2 & 0.88 & 4.77 & 1.10 \\
         $\mathrm{Qwen2.5\textrm{-}7B_{NO\_STAGE2}}$ & 24.4 & \multicolumn{2}{c}{4.1} & 20.9 & 20.7 & 3.3 & 20.2 & 0.88 & 4.66 & 1.24 \\
         $\mathrm{Qwen2.5\textrm{-}7B_{ALL}}$ & \textbf{26.4} & \multicolumn{2}{c}{\textbf{4.6}} & \textbf{22.8} & \textbf{22.7} & \textbf{3.7} & \textbf{21.1} & \textbf{0.88} & \textbf{4.75} & \textbf{1.28} \\
         \bottomrule
    \end{tabular}
    \caption{Averaged Evaluation Metrics (\%) and GPT-4 assessment (\textit{Flu.} is fluency and \textit{Ade.} is adequacy) of text generated from EEG signals using different LLMs. A comparison is made between chance-level performance (with only object label given as input ($\mathrm{ONLY\_OBJ}$), and the object label and a random embedding given as input ($\mathrm{OBJ+RAND\_EMB}$) and only EEG embeddings given as input ($\mathrm{ONLY\_EEG}$) and our solutions without Stage 2 ($\mathrm{NO\_STAGE2}$), and the complete solution with all stages ($\mathrm{ALL}$).}.
    \label{tab:results}
\end{table*}

\section{Results}
\label{sec:results}

\begin{figure*}[t]
    \centering
    \begin{subfigure}[b]{0.32\textwidth}
        \includegraphics[width=\textwidth]{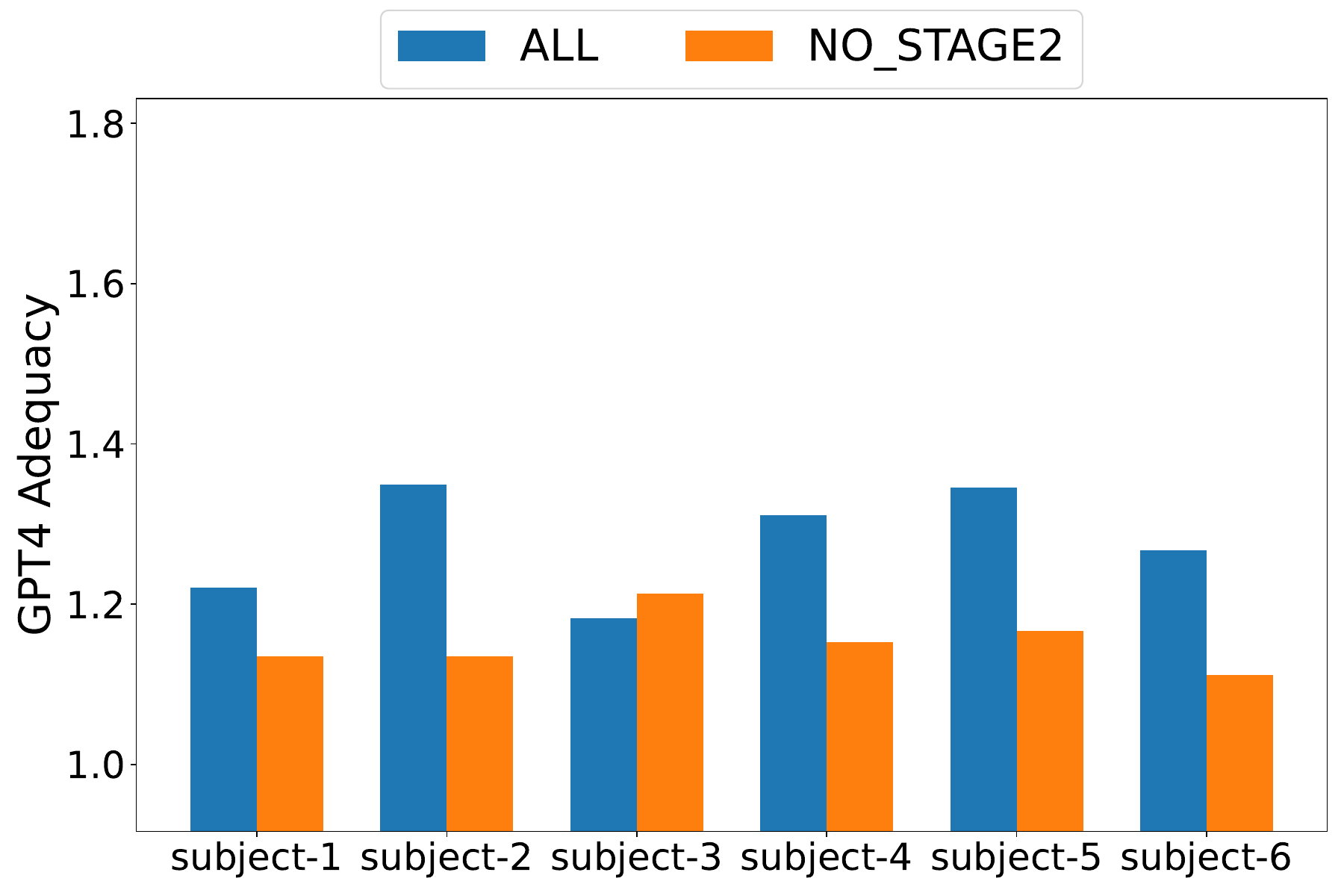}
        \caption{Llama-3-8B - GPT4 Adequacy}
    \end{subfigure}
    \hfill
    \begin{subfigure}[b]{0.32\textwidth}
        \includegraphics[width=\textwidth]{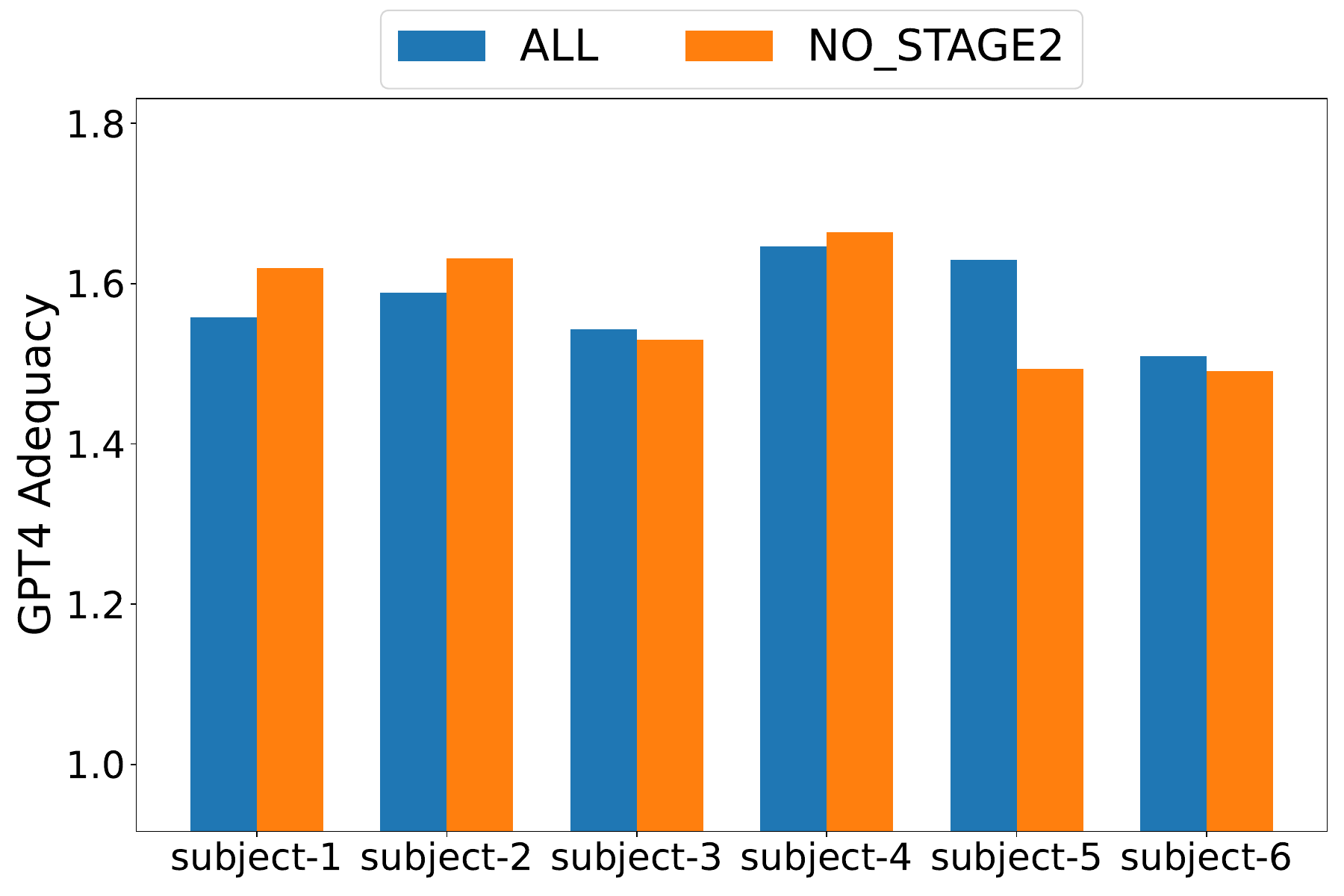}
        \caption{Mistral-7B-v0.3 - GPT4 Adequacy}
    \end{subfigure}
    \hfill
    \begin{subfigure}[b]{0.32\textwidth}
        \includegraphics[width=\textwidth]{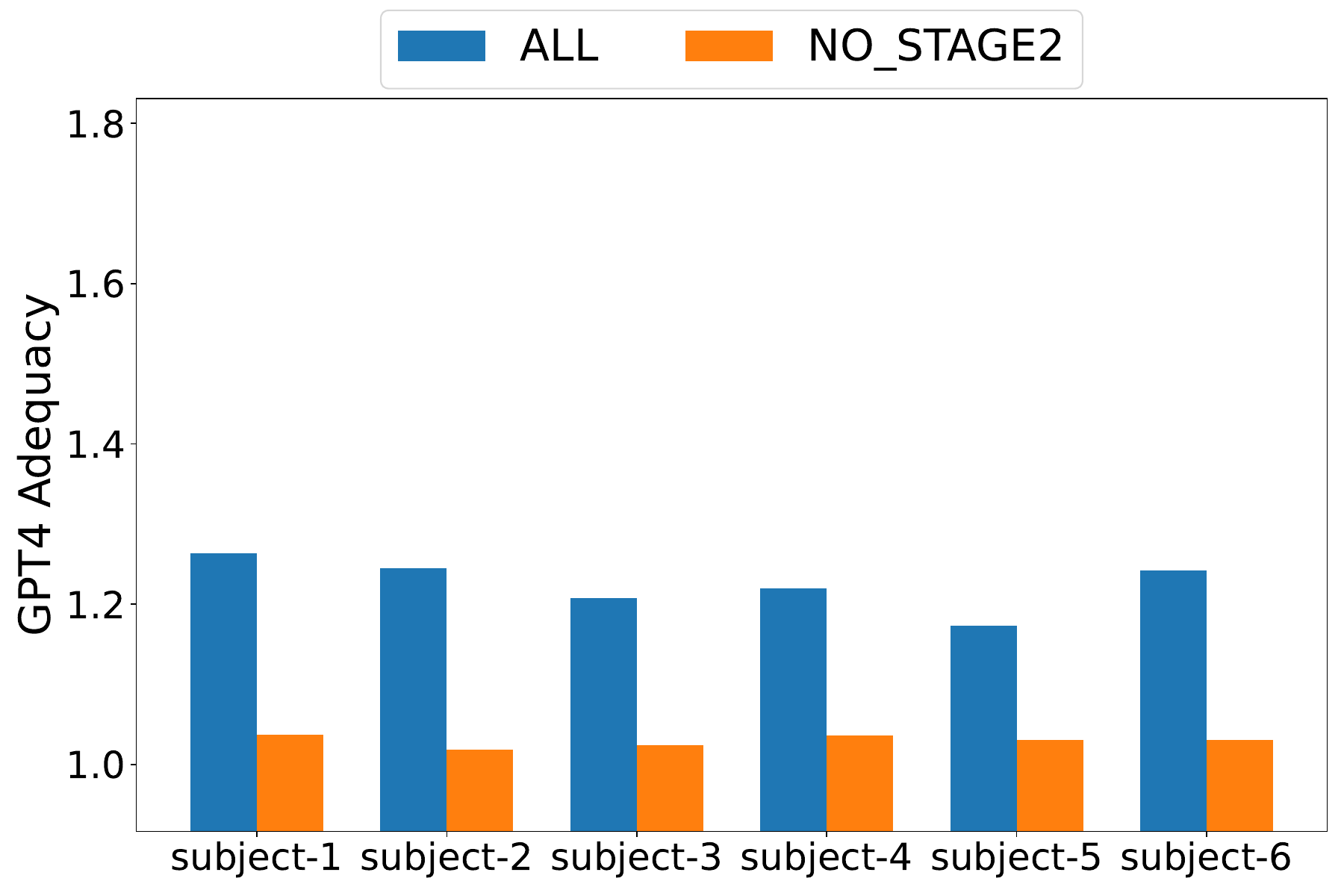}
        \caption{Qwen2.5-7B - GPT4 Adequacy}
    \end{subfigure}

    \vspace{0.5cm} 

    \begin{subfigure}[b]{0.32\textwidth}
        \includegraphics[width=\textwidth]{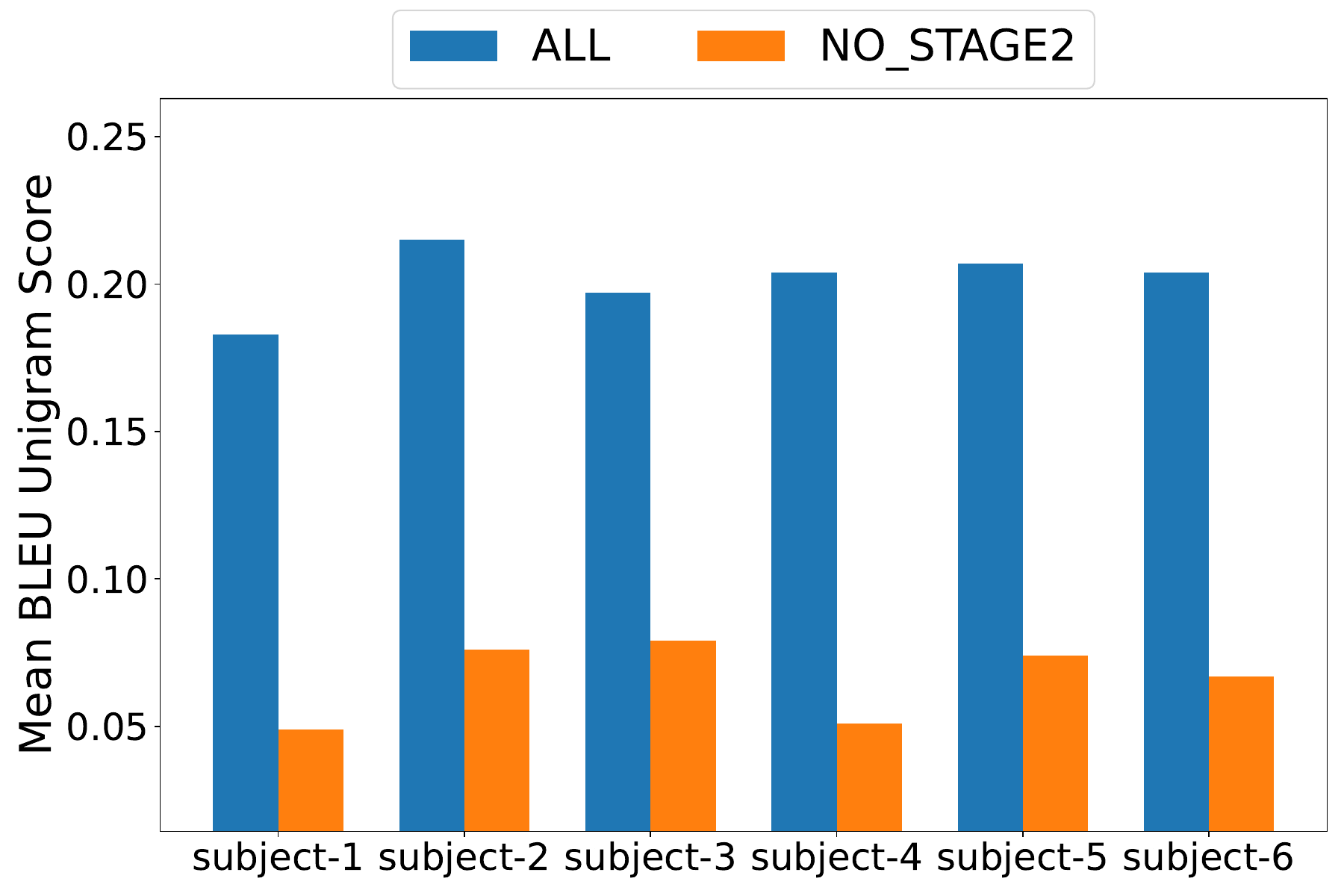}
        \caption{Llama-3-8B - BLEU Unigram}
    \end{subfigure}
    \hfill
    \begin{subfigure}[b]{0.32\textwidth}
        \includegraphics[width=\textwidth]{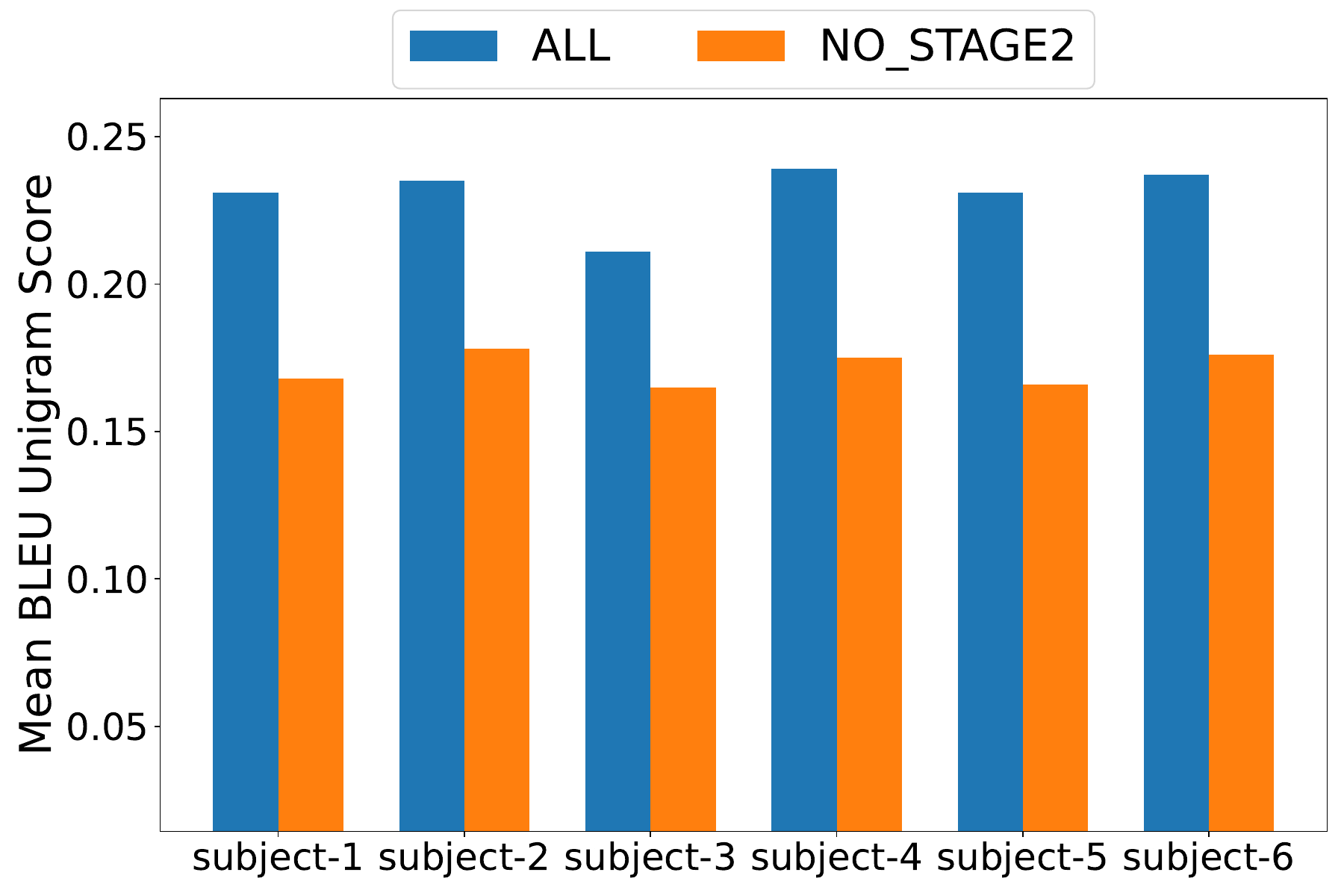}
        \caption{Mistral-7B-v0.3 - BLEU Unigram}
    \end{subfigure}
    \hfill
    \begin{subfigure}[b]{0.32\textwidth}
        \includegraphics[width=\textwidth]{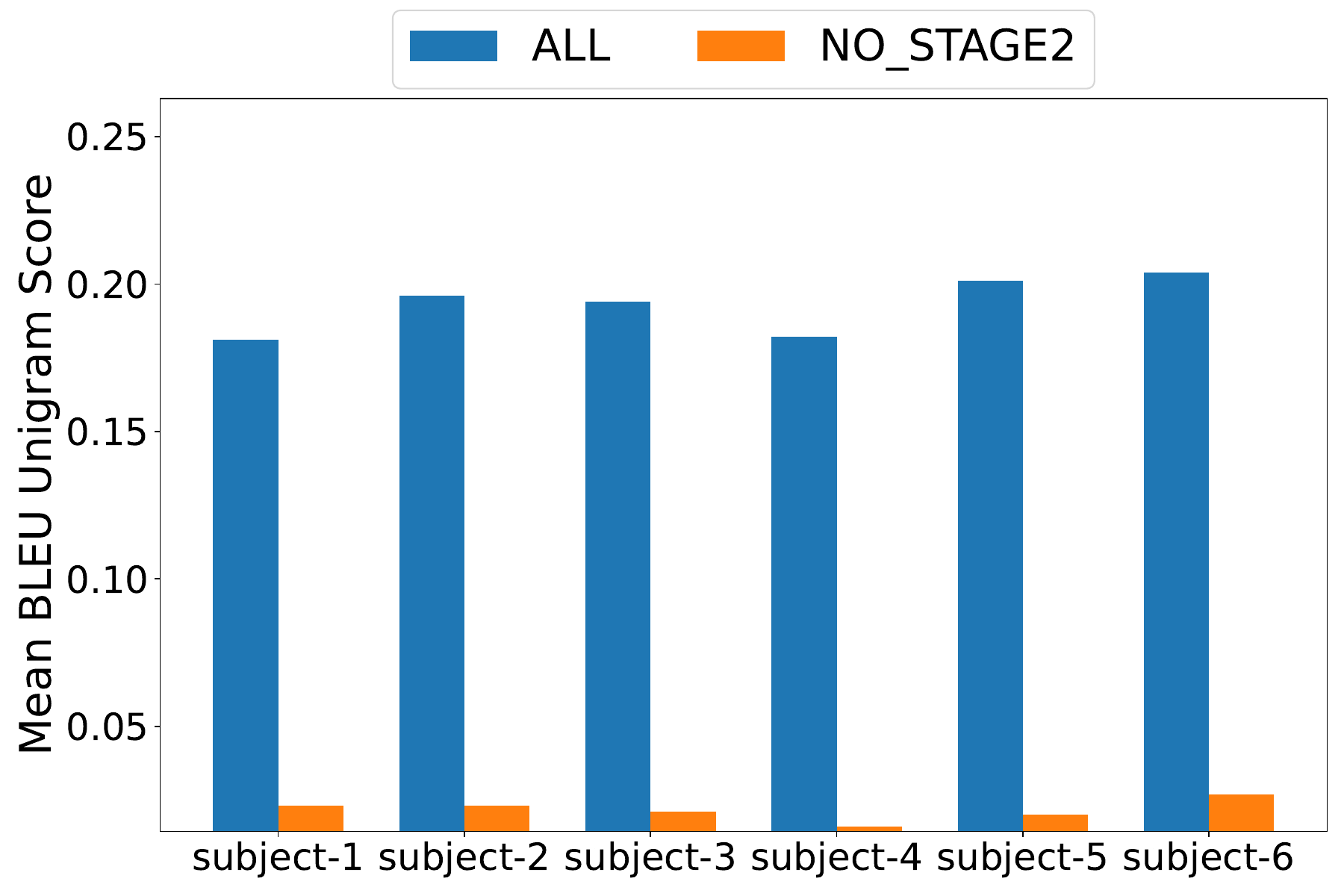}
        \caption{Qwen2.5-7B - BLEU Unigram}
    \end{subfigure}

    \caption{Subjectwise analysis comparing $\mathrm{ALL}$ and $\mathrm{NO\_STAGE2}$ variants across subjects for GPT-4 Adequacy and BLEU Unigram metrics, evaluated using different models. For BLEU scores, the $\mathrm{ALL}$ variants show a noticeable improvement across all six subjects compared to the $\mathrm{NO\_STAGE2}$ variants. Although numerically smaller, a consistent improvement in Adequacy is also observed with the $\mathrm{ALL}$ variants, which is significant in context of noisy EEG data.}
    \label{fig:subjectwise}
\end{figure*}

Table \ref{tab:results} presents a comprehensive comparison of metrics across different models and setups. From these results, it is evident that the complete approach, denoted as \textbf{ALL}, consistently outperforms other setups across all evaluation metrics. As expected, chance-based baselines like $\mathrm{ONLY\_OBJ}$ and $\mathrm{OBJ+RAND\_EMB}$ exhibit poor performance. With our proposed methodology, the LLaMA-v3-8B\_ALL model achieved a \textbf{BLEU-N (N=1)} score of 25.5\%, Mistral-7B\_ALL scored 26\%, and Qwen2.5-7B\_ALL reached 22.7\%—-all significantly higher than their respective chance scores under the chance setups. In particular, some models, mainly the LLaMa variants, show increased sensitivity to random input, leading to further reduction in the scores with chance setups. 

We will first discuss the inferences from the results obtained with other alternative setups before delving into a subject-wise analysis.

\subsection{Comparison with Stage 2 Omission (NO\_STAGE2)}
The \textbf{ROUGE-N (N=1)} score for LLaMA3-8B's $\mathrm{ALL}$ variant is 30.0\%, whereas the $\mathrm{NO\_STAGE2}$ variant achieves only 26.9\%, indicating a significant improvement when Stage 2 is added. The same trend is observed for other metrics like BLEU, METEOR, and BERT Score, and is observed across other models as well. In terms of adequacy as measured by GPT-4, the $\mathrm{ALL}$ variant stands out except for Qwen model which shows a tendency to copy the object and produce shorter sentences in the case of $\mathrm{ONLY\_OBJ}$ which is positively rated by GPT-4. 
Overall, the table demonstrates that incorporating Stage 2 (as detailed in Section \ref{sec:method}) -- which aligns EEG embeddings with image embeddings using a CLIP-based supervision strategy -- contributes to higher-quality text generation.

\subsection{Generation performance without object labels in the input}
While our initial hypothesis was that EEG embeddings -- derived from noisy multichannel EEG data -- might not fully capture complex thoughts, thus requiring additional inputs like object labels, it is pleasantly surprising to observe that even without object labels, the models ($\mathrm{ONLY\_EEG}$) perform comparably with the best models ($\mathrm{ALL}$). This underscores the effectiveness of aligning EEG embeddings with vision embeddings in stage 1 and pretraining the LLM, particularly the projectors, in stage 2 with vision embeddings.

\subsection{Subject-wise Analysis}
For this analysis, each subject’s EEG data is used to independently train and test the LLMs, simulating a personalized solution. The dataset comprises six subjects, allowing for individual analysis to evaluate the robustness of the approach across different participants. As depicted in Figure \ref{fig:subjectwise}, in the subject-wise analysis, the advantages of the complete approach (ALL) become even more prominent. Cross-subject and in-subject experiments consistently favor the ALL configuration, with significant improvements in \textit{Adequacy} score when Stage 2 is included, especially for LLaMA3-8B and Qwen2.5-7B models. These improvements, though numerically small, are crucial in the context of EEG data where every bit of alignment and finetuning matters due to its inherently noisy and sparse nature. 

The $\mathrm{NO\_STAGE2}$ configuration, which omits the essential alignment step between EEG embeddings and image embeddings, consistently demonstrates lower performance across subjects, as illustrated by the BLEU Unigram scores in Figure \ref{fig:subjectwise}. This validates our hypothesis that direct fine-tuning without the warm-up provided by Stage 1 and Stage 2 is insufficient for EEG data. The challenge is further exacerbated by the inherent difficulties associated with EEG data, which, even when ethically collected and cleaned, still suffer from limited data availability and significant variability across sessions and subjects.

Subject-wise analysis is essential in practice due to the inherently private and sensitive nature of EEG data. Models like ours, designed for thought-to-text translation, must be developed and deployed within privacy-preserving settings. Creating a personalized solution without access to extensive cross-subject EEG data can be challenging. To this end, our multi-stage approach demonstrates that pretraining on non-EEG data (such as images) and fine-tuning on small amounts of subject-specific EEG data opens new avenues for privacy-preserving personalized EEG-LLM model development.

\subsection{Qualitative Inspection and Basic Error Analysis}
For qualitative inspection, we compared the generated descriptions of images produced by both the GPT-4 and the \textsc{Mistral-7B-v0.3} model (ALL and $\mathrm{ONLY\_EEG}$ variants). Table \ref{tab:qualitative_anecdotal_results} in the Appendix section presents notable examples from our assessment of the model's generated descriptions. In many instances, the approach with input of EEG + predicted object label generates highly accurate descriptions, as illustrated in positive examples 2 and 4 for the piano and pumpkin, respectively. However, in cases of misgeneration, we encounter not only significantly inaccurate outputs and hallucinations, as seen in examples 6, 7, and 8, but also instances of genuine confusion. For example, the EEG signal for a flower is misidentified as a mushroom in example 5, leading to incorrect generation. This misidentification indicates potential areas for improvement in object classification.

However, a key advantage of the EEG-only approach is that even when the predicted object label is incorrect, the model can still produce reasonably coherent descriptions, as seen in anecdotal example 1. This highlights the robustness of the EEG embeddings in guiding the language model's generation. Similar observations were made with other models based on LLaMa and Qwen architectures, further validating the consistency and reliability of our approach across various LLM frameworks. We acknowledge that the selected examples are purposefully chosen to illustrate different scenarios. 
\section{Conclusion and Future Work}
\label{sec:conclusion}
In this paper, we introduced a novel approach to convert EEG signals into text, leveraging instruction-tuned LLMs fine-tuned with EEG data. Our method progresses through three stages: training an EEG encoder for feature extraction, fine-tuning LLMs on multimodal data, and further refining them with EEG embeddings for direct text generation from neural signals. Validation on a public EEG dataset 
demonstrates the efficacy of popular LLMs in "transforming" thoughts evoked by viewing images into  text. We observed significant performance improvements compared to chance evaluation, and our methodology, incorporating all stages, performed well in both cross-subject and in-subject analyses, as validated through quantitative evaluation. The qualitative evaluation provided further insights into various scenarios involving EEG signals and object labels versus EEG signals alone as inputs for generating text. These evaluations not only reinforce the effectiveness of our methodology for efficient text generation but also underscore the potential of utilizing EEG data alone to achieve the desired results. However, instances of misidentification that result in incorrect outputs reveal opportunities for further improvements in text generation. 

Our future work will focus on optimizing the model architecture, leveraging foundational pretrained EEG models on diverse dataset, improving EEG-text alignment through stage 2 training on large scale image datasets and diverse tasks such as optical character recognition, question answering and summarizing and exploring practical applications in healthcare and assistive technologies, marking a significant stride toward accessible "thoughts-to-text" systems.

\section*{Limitations}

Extracting fine-grained information from EEG signals presents challenges due to high data-to-noise ratio and low spatial resolution. Despite these difficulties, EEG signals can still identify object categories, which can then be used with a generic prompt to aid in text generation. However, misclassification of similar-shaped objects (see the Appendix, Figure \ref{fig:confusion_matrix}), such as mistaking a mushroom for a flower, underscores potential ambiguities in object recognition. In addition, there were instances where the encoder generated unrelated descriptions, such as identifying a coffee maker as a computer or an elephant as a panda. Object classification accuracy varies among subjects (see the Appendix, Table \ref{tab:subject_wise_accuracy}) due to individual differences in interpreting images, leading to diverse EEG signal variations and increased prediction variance. One approach to address this is by training personalized models for each subject and assessing their performance. Additionally, implementing methods that enhance the generalizability of predictions across subjects could be explored. Another challenge is data scarcity; Deep Learning models typically require substantial data for training. Hence, acquiring more high-quality, multi-channel EEG data under controlled experimental conditions is crucial to reduce noise. Despite these challenges, our quantitative and qualitative findings demonstrate promising results. We believe that additional training on larger datasets and rigorous controlled experiments will significantly improve performance.

While reading thoughts can be beneficial for individuals with limited ability to communicate, the major risk lies in the potential misuse of BCIs to intrude into thoughts without consent. However, with appropriate measures and regulations, these risks should not hinder advancements in understanding and translating human cognition, as the benefits outweigh the challenges.

\section*{Ethics Statement}
For this work, we utilized anonymized open-source EEG data, acknowledging the sensitivity of EEG data and the imperative of ethical compliance. All experimental data used in our research were anonymized to protect participant privacy and uphold ethical standards. Additionally, we employed OpenAI's ChatGPT-4 system to enhance writing efficiency by generating LaTeX code, ensuring concise sentences, and aiding in error debugging.

\bibliography{nacl2025}
\bibliographystyle{acl_natbib}
\clearpage
\onecolumn

\section{Appendix}
\begin{figure*}
\vspace{-2.5cm}
     \centering
    \includegraphics[width=1\textwidth]{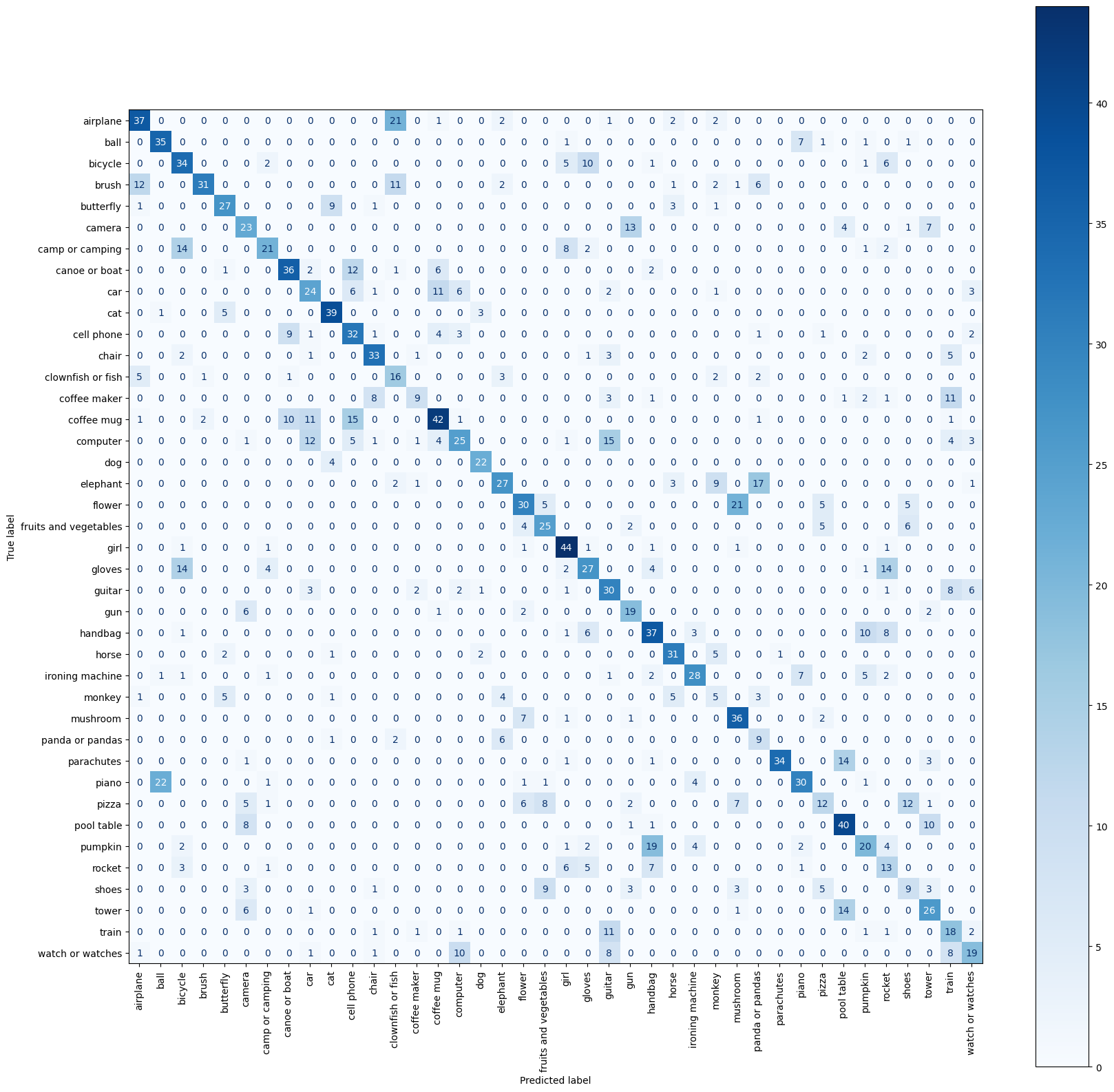}
   \caption{Confusion Matrix for classification results from stage 1. Some misclassifications are flowers being identified as mushrooms, airplane as identified as clownfish or fish, elephant as a panda etc.} 
	\label{fig:confusion_matrix}
\end{figure*}

We present supporting anecdotal examples in Table \ref{tab:qualitative_anecdotal_results}, subject-wise classification results in Table \ref{tab:subject_wise_accuracy}, and image classification results in Figure \ref{fig:confusion_matrix}.

\begin{table*}[h]
    \centering
    \small
    \begin{tabular} 
    {>{\centering\arraybackslash}p{0.01\linewidth}>{\centering\arraybackslash}p{0.2\linewidth}>{\centering\arraybackslash}p{0.05\linewidth}>{\centering\arraybackslash}p{0.1\linewidth}>{\centering\arraybackslash}p{0.1\linewidth}>{\centering\arraybackslash}p{0.15\linewidth}>{\centering\arraybackslash}p{0.15\linewidth}}
    \toprule
          \textbf{ID}&\textbf{Image} & \textbf{Input}&\textbf{Reference Object} &\textbf{Predicted Object }&  \textbf{Reference  Description}& \textbf{Thought2Text Description} \\
          \midrule
  1&\raisebox{-1cm}{\includegraphics[width=0.15\textwidth, height=15mm]{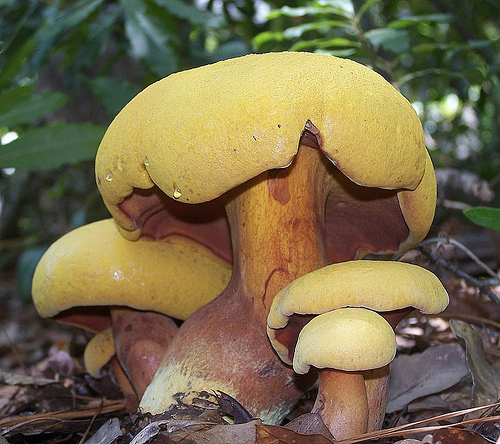}}& EEG& mushroom& flower&  A large yellow mushroom with a brown stem and a brown cap, surrounded by green foliage. 
&\textcolor{OliveGreen}{A group of mushrooms growing on a log.}\\
         \midrule
          2&\raisebox{-1cm}{\includegraphics[width=0.15\textwidth, height=15mm]{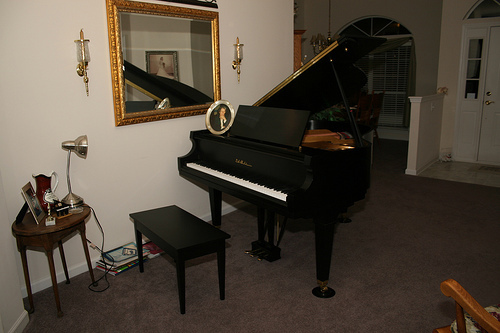}} & EEG + OBJ& piano&piano&  A black grand piano in a living room.  & \textcolor{OliveGreen}{A grand piano with a stool in front of it.} \\
         \midrule
  3&\raisebox{-1cm}{\includegraphics[width=0.15\textwidth, height=15mm]{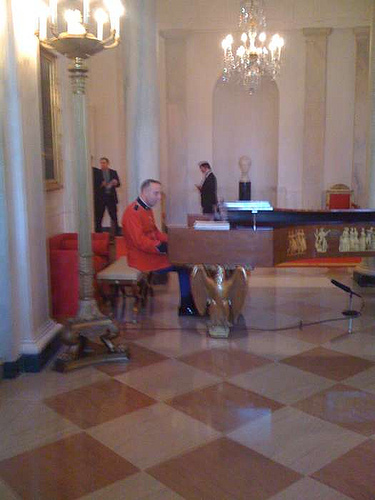}} & EEG& piano& piano&  A man in a red coat and black pants is playing a piano in a room with a chandelier. 
&\textcolor{OliveGreen}{A man is playing the piano in a dimly lit room.}
\\
         \midrule
          4&\raisebox{-1cm}{\includegraphics[width=0.15\textwidth, height=15mm]{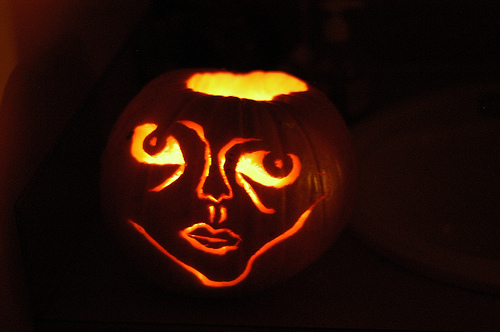}} & EEG + OBJ& pumpkin&pumpkin& A carved pumpkin with a face and eyes, sitting on a table.  & \textcolor{OliveGreen}{A carved pumpkin with a spooky face on it.} \\
         \midrule
  5&\raisebox{-1cm}{\includegraphics[width=0.15\textwidth, height=15mm]{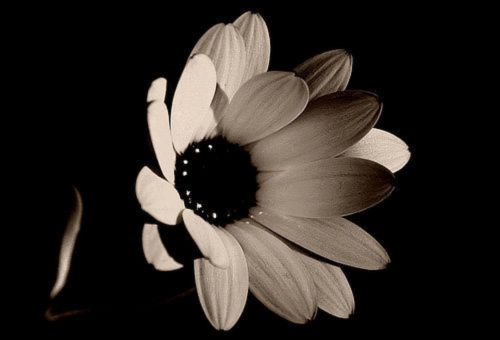}} & EEG + OBJ& flower&mushroom& A black and white photograph of a single daisy with a white center and a dark brown center.&{\textcolor{BrickRed}{A group of mushrooms growing on a log.}}\\
 \midrule
  6&\raisebox{-1cm}{\includegraphics[width=0.15\textwidth, height=15mm]{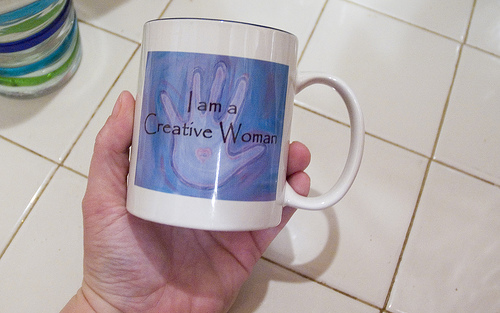}} & EEG + OBJ& coffee mug&coffee mug& A hand holding a mug with a blue background and a handprint design. &{\textcolor{BrickRed}{A person holding a coffee mug with the words "World's Best Dad" written on it.}}\\
  \midrule
    7&\raisebox{-1cm}{\includegraphics[width=0.15\textwidth, height=15mm]{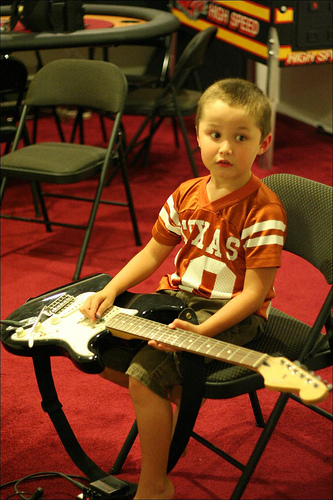}} & EEG & guitar&watch or watches&  A young boy sitting on a chair playing a guitar.  &{\textcolor{BrickRed}{A man is holding a guitar in front of a microphone.}}\\
  \midrule
  8&\raisebox{-1cm}{\includegraphics[width=0.15\textwidth, height=15mm]{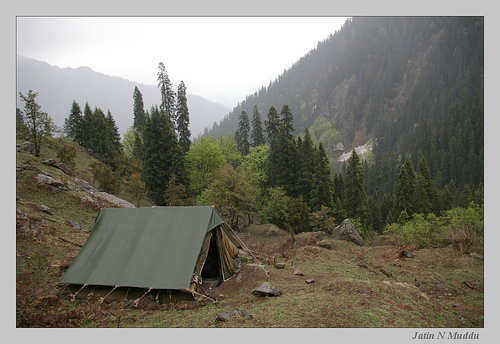}} & EEG + OBJ& camp or camping&camp or camping& A tent in a mountainous area with trees and fog. &{\textcolor{BrickRed}{A tent set up in a forest with a campfire nearby.}}\\
 \bottomrule
    \end{tabular}
    \caption{Sample positive (in green) and negative (in red) anecdotal examples using the \textsc{Mistral-7B-v0.3} $ALL$ and $EEG\_ONLY$ variants that take different inputs: EEG signals + object information and EEG signals alone.}
    \label{tab:qualitative_anecdotal_results}
\end{table*}

\begin{table*}
\vspace{-1.5cm}
\centering
\footnotesize
\begin{tabular}{p{2cm}p{2cm}p{1.8cm}}
\toprule
\textbf{Subject} & \textbf{Accuracy} \\    
\midrule
Subject 1 & 0.631 \\
Subject 2 & 0.631 \\
Subject 3 & 0.544 \\
Subject 4 & 0.600 \\
Subject 5 & 0.594 \\
Subject 6 & 0.531 \\
\cmidrule{1-2}
Overall  & 0.530 \\ 
\bottomrule
\end{tabular}
\caption{Subject-wise Object Classification Accuracy obtained using raw EEG Signals from Stage-1 (\%). The train-test splits used to obtain these results are identical to those used in the main experiments.}
\label{tab:subject_wise_accuracy}
\end{table*}

\end{document}